%% file: syntactic_lm.tex
\documentclass[11pt,a4paper]{article}
\usepackage[hyperref]{emnlp-ijcnlp-2019}
\usepackage{etex}  
\usepackage{standalone}
\usepackage{amsmath}
\usepackage{amssymb}
\usepackage{amsfonts}
\usepackage{graphicx}
\usepackage{times}
\usepackage{latexsym}
\usepackage{mathtools}
\usepackage{soul}
\usepackage{graphicx}
\usepackage{float}
\usepackage{booktabs}
\usepackage{xspace}
\usepackage{url}
\usepackage{color}
\usepackage[skip=5pt]{caption}
\usepackage{subcaption}
\usepackage{array}
\usepackage{tabu}
\usepackage{makecell}
\usepackage{subcaption}
\usepackage{diagbox}
\usepackage{enumitem}
\usepackage{multirow}
\usepackage{verbatim}
\usepackage{tabulary}
\usepackage{booktabs}
\usepackage[mathscr]{euscript}
\usepackage{mathtools}
\usepackage{amssymb}
\usepackage{algorithm}
\usepackage{algpseudocode}
\algrenewcommand{\algorithmiccomment}[1]{\leavevmode$\triangleright$ #1}
\usepackage{stmaryrd}
\usepackage{amsthm}
\usepackage{tikz-dependency}
\usetikzlibrary{automata,decorations.markings,arrows,positioning,matrix,calc,patterns,angles,quotes,calc}
\usepackage{adjustbox}
\usepackage[toc,page]{appendix}
\usepackage{tabularx}
\usepackage[shortcuts]{extdash} 
\usepackage{paralist}

\usepackage{bm}
\definecolor{orange}{rgb}{1,0.5,0}
\definecolor{mdgreen}{rgb}{0.05,0.6,0.05}
\definecolor{mdblue}{rgb}{0,0,0.7}
\definecolor{dkblue}{rgb}{0,0,0.5}
\definecolor{dkgray}{rgb}{0.3,0.3,0.3}
\definecolor{slate}{rgb}{0.25,0.25,0.4}
\definecolor{gray}{rgb}{0.5,0.5,0.5}
\definecolor{ltgray}{rgb}{0.7,0.7,0.7}
\definecolor{purple}{rgb}{0.7,0,1.0}
\definecolor{lavender}{rgb}{0.65,0.55,1.0}

\aclfinalcopy 

\newcommand{\appref}[1]{Appendix~\ref{app:#1}}

\newcommand{\secref}[1]{\S\ref{sec:#1}}
\newcommand{\figref}[1]{Fig.~\ref{fig:#1}}
\newcommand{\com}[1]{}
\newcommand{\camready}[1]{}
\newcommand{\resolved}[1]{}
\newcommand{\modelname}[0]{{PaLM}\xspace} 

\title{\modelname: A Hybrid Parser and Language Model}

\newcommand{\ensuretext}[1]{#1}
\newcommand{\marker}[2]{\ensuremath{^{\textsc{#1}}_{\textsc{#2}}}}

\newcommand{\arkcomment}[3]{\ensuretext{\textcolor{#3}{[#1 #2]}}}

\newcommand{\nascomment}[1]{\arkcomment{\marker{NA}{S}}{#1}{blue}}
\newcommand{\hao}[1]{\arkcomment{\marker{H}{P}}{#1}{mdgreen}}

\newcommand{\isection}[2]{\section{#1}\label{sec:#2}}
\newcommand{\isubsection}[2]{\subsection{#1}\label{sec:#2}}
\newcommand{\argmax}[1]{\underset{#1}{\operatorname{argmax}}\;}

\newcommand{\interalia}[1]{\citep[\emph{inter alia}]{#1}}

\DeclareCaptionFont{tiny}{\tiny}

\def\va{{\mathbf{a}}}

\def\vc{{\mathbf{c}}}

\def\vf{{\mathbf{f}}}
\def\vg{{\mathbf{g}}}
\def\vh{{\mathbf{h}}}

\def\vu{{\mathbf{u}}}

\def\vx{{\mathbf{x}}}
\def\vy{{\mathbf{y}}}
\def\vz{{\mathbf{z}}}

\def\mW{{\mathbf{W}}}

\def\gH{{\mathcal{H}}}

\def\gL{{\mathcal{L}}}

\def\gO{{\mathcal{O}}}

\newcommand{\algoref}[1]{Alg.~\ref{algo:#1}}
\newcommand{\equref}[1]{Eq.~\ref{eq:#1}}
\newcommand{\repo}[0]{\url{https://github.com/Noahs-ARK/PaLM}}

\author{Hao Peng$^\spadesuit$ \quad Roy Schwartz$^\spadesuit$$^\diamondsuit$ \quad
	Noah A. Smith$^\spadesuit$$^\diamondsuit$ \\
	$^\spadesuit$Paul G. Allen School of Computer Science \& Engineering,
	University of Washington, Seattle, WA, USA \\
	$^\diamondsuit$Allen Institute for Artificial Intelligence, Seattle, WA, USA \\
	{\tt \{hapeng,roysch,nasmith\}@cs.washington.edu}}

\date{}

\begin{document}
\maketitle

\input{text/abstract}
\input{text/intro}

\input{text/model}

\input{text/experiments}
\input{text/conclusion}

\clearpage

\bibliography{syntactic_lm}
\bibliographystyle{acl_natbib}

\clearpage
\input{text/appendix}

\end{document}

%% file: text/abstract.tex
\begin{abstract}
We present~\modelname, a hybrid \textbf{pa}rser and neural \textbf{l}anguage \textbf{m}odel.
Building on an RNN language model, 
\modelname adds an attention layer over \emph{text spans} in the left context.
An unsupervised constituency parser can be derived from its attention weights,
using a greedy decoding algorithm\com{~\citep{Stern:2017}}.
We evaluate \modelname~on language modeling, 
and empirically show that it
outperforms strong baselines.
If syntactic annotations \emph{are} available, 
the attention component can be trained in a supervised manner, 
providing syntactically-informed representations of the context, and further improving language modeling performance.



\end{abstract}

%% file: text/intro.tex
\isection{Introduction}{intro}

Recent language models have shown very strong 
data-fitting performance~\citep{Jozefowicz:2016,Merity:2018}.
They offer useful products including, most notably, contextual embeddings~\citep{Peters:2018,Radford:2019}, 
which benefit many NLP tasks such as 
text classification~\cite{Howard:2018} and dataset creation~\citep{Zellers:2018}.

Language models are typically trained on large amounts of raw text, 
and therefore do not explicitly encode any notion of structural information. 
Structures in the form of syntactic trees have been shown to benefit both
classical NLP models~\interalia{Gildea:2002,Punyakanok:2008,Das:2012}
and recent state-of-the-art neural models~\interalia{dyer2016rnng,Swayamdipta:2018,peng2018backprop,Strubell:2018}. 
In this paper we show that LMs can benefit
from syntactically-inspired encoding of the context.

\begin{figure}[!t]
\includegraphics[width=.98\linewidth]{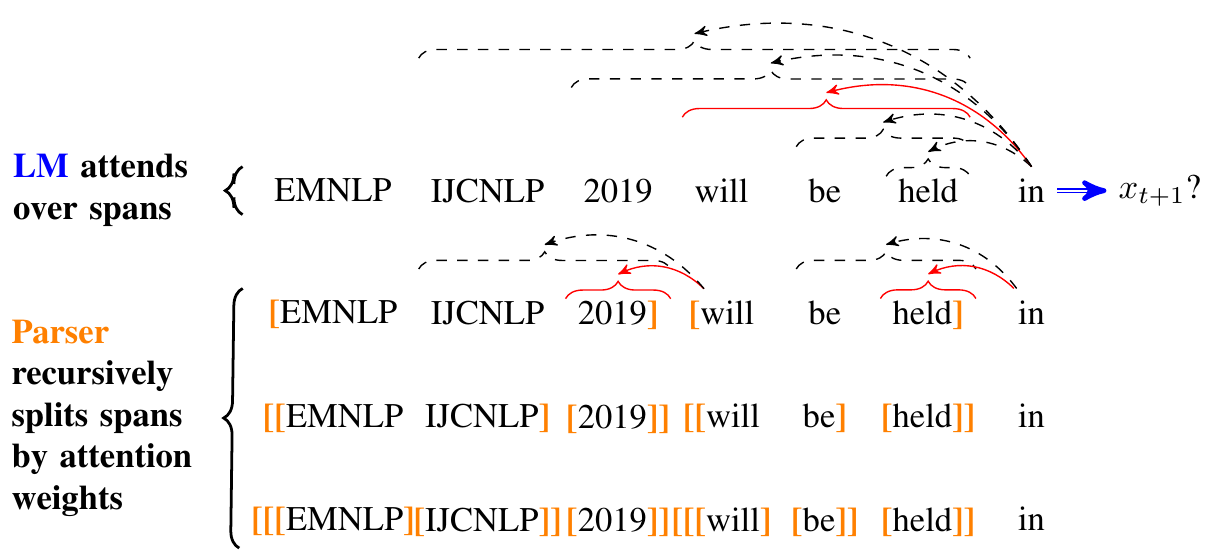}
\caption{\label{fig:example} An illustration of~\modelname. The LM (first line) predicts the next word  ($x_{t+1}$, double blue arrow) by attending over previous spans ending in time $t-1$ (dashed lines). 
The parser (lines 2--4) splits the prefix into two spans (line 2) by the taking the top scoring attended span (red solid line) and the prefix leading to it.
It then recursively splits the two sub-spans using the same procedure
(line 3). 
Finally, spans of length two are trivially split into terminal nodes
(line 4). }
\end{figure}

We introduce \modelname~(\textbf{pa}rser and \textbf{l}anguage \textbf{m}odel; \figref{example}), 
a novel hybrid model combining an RNN language model with a constituency parser. 
The LM in \modelname~attends over \emph{spans of tokens},
implicitly learning which syntactic constituents are likely.
A span-based parser is then derived from the attention information~\citep{Stern:2017}.


\modelname~has several benefits. 
First, it is an intuitive and lightweight way of incorporating 
structural information (\S\ref{sec:span}),
requiring no marginal inference, which can be computationally expensive~\interalia{jelinek1991computation,chelba1998exploiting,roark2001probabilistic,dyer2016rnng,buys2018neural,kim2019unsupervised}.
Second, the attention can be 
syntactically informed,
in the sense that the attention component
can optionally be supervised using syntactic annotations,
either through pretraining
or by joint training with the LM (\S\ref{sec:span_parsing}).
Last, \modelname~can derive an unsupervised constituency parser~(\S\ref{sec:span_parsing}),
whose parameters are estimated purely 
using the language modeling objective.

To demonstrate the empirical benefits of \modelname, 
we experiment with language modeling (\S\ref{sec:experiments}).
\modelname outperforms the AWD-LSTM model \cite{Merity:2018} 
on both the Penn Treebank (PTB;~\citealp{Marcus:1993}) 
and WikiText-2~\citep{Merity:2017} datasets by small but consistent margins in the unsupervised setup. 
When the parser is trained jointly with the language model, 
we see additional perplexity reductions in both cases.  
Our implementation is available at~\repo.

%% file: text/model.tex
\isection{\modelname---Parser and Language Model}{method}
We describe \modelname in detail.
At its core is an attention component,
gathering the representations of preceding \emph{spans}
at each time step.
Similar to self-attention,
\modelname~can be implemented on top of
RNN encoders~\citep{parikh2016decomposable}, or as it is~\citep{vaswani2017attention}.
Here we encode the tokens using a left-to-right RNN, denoted with vectors $\vh_t$.\footnote{We experiment with a strong LSTM implementation for language modeling~\citep{Merity:2018}, see \secref{experiments}.}

Below we describe the span-attention component and the parsing algorithm. 
We use $[i, j], i\leq j$ to denote text span $x_i\dots x_j$, 
i.e., inclusive on both sides. 
When $i=j$, it consists of a single token.

\com{
\paragraph{Notations and background.}
A language model learns a joint distribution of 
a corpus, which typically autoregressively factorizes as
\begin{align}
\mathbb{P}_\theta(\vx) = \prod_{t=1}^{n}\mathbb{P}_\theta(x_t\mid \vx_{<t}),
\end{align}
where $\vx_{<t}=x_1\dots x_{t-1}$ 
denotes the preceding tokens up to time step $t$ (exclusive),
and $\theta$ denotes the model parameters.

Recurrent neural networks provide natural solutions
to encoding variable-length inputs,
and prove successful in modeling languages~\interalia{zaremba2014rnn,gal2016theoretically,Merity:2018}.
At a very high level, a recurrent cell
can be seen as a parameterized function,
which is recessively applied along the sequence:
\begin{align}
	\vh_{t} = \vf\left(\vh_{t-1}, \vz_t\right),
\end{align}
where $\vz_t$ denotes the embedding vector of token $x_t$,
and $\vh_{t}$ vectors the encoded hidden states, which are 
used for onward computation, e.g.,
feeding into the next RNN layer or into a classifier.

Such architectures are sometimes augmented with self attention mechanisms,
allowing for better modeling of long-range dependencies~\citep{parikh2016decomposable}.
An attention function gathers the context, usually with a 
(normalized) weighted sum, where the weights are computed 
from the current hidden state and the context.
A language model is constrained from 
conditioning on future tokens, and hence
its attention function can only access $\vh_{<t}$
at timestep $t$.
 }

\isubsection{Span Attention}{span}

We want the language model attention to gather context information
aware of syntactic structures.
A constituency parse can be seen as a collection of syntactic constituents, i.e.,  token spans.
Therefore we attend over preceding spans\com{, which we describe in this section}.\footnote{Standard token-based self-attention
	naturally relates to dependency structures through head selection~\citep{Strubell:2018}.
	In a left-to-right factored language model, dependencies are
        less natural if we want to allow a child to precede its
        parent.}

At step $t$, \modelname~attends over the spans ending at $t-1$, up to a
maximum length $m$, i.e., $\{[i, t-1]\}_{i=t-m}^{t-1}$.\footnote{$m$
  is set to 20.
	This reduces the number of considered spans from~$\gO{(n^2)}$
	to~$\gO{(mn)}$. 
	Besides practical concerns,
	it makes less sense if a phrase goes
	beyond one single sentence (the average sentence length of WSJ training sentences is 21).
	}
Essentially, this can be seen as
splitting the prefix span $[t-m, t-1]$ into two,
and attending over the one on the right.
Such a span attention mechanism
is inspired by the top-down greedy span parser of~\citet{Stern:2017},
which recursively divides phrases.
In \secref{span_parsing},
we will use a similar algorithm
to derive a constituency parser from the span attention weights.

\paragraph{Bidirectional span representation with rational RNNs.}
Meaningful span representations
are crucial in span-based tasks~\interalia{Lee:2017,peng2018learning,Swayamdipta:2018}.
Typical design choices are based on
start and end token vectors contextualized by
bidirectional RNNs.
However, a language model does not have access to future words,
and hence running a backward RNN from right to left is less straightforward:
one will have to start
an RNN running at each token,
which is computationally daunting~\citep{kong2015segrnn}.
To compute span representations efficiently, we use \emph{rational} RNNs~(RRNNs; \citealp{Peng:2018}).

RRNNs are a family of RNN models, where the recurrent function 
can be computed with weighted finite-state automata (WFSAs).
We use the unigram WFSA--inspired 
 RRNN~\citep{Peng:2018},  where the cell state update is
\begin{subequations}\label{eq:rrnn}
	\begin{align}
	\vf_t &=\bm{\sigma}\left(\mW_f\vh_t\right),\\
	\vu_t &= (\mathbf{1} - \vf_t)\odot\tanh\left(\mW_u\vh_t\right),\\
	\vc_{t} &= \vf_{t} \odot \vc_{t-1} + \vu_{t}.\label{eq:rrnn:c}
	\end{align}
\end{subequations}
$\vf_t$ is a forget gate implemented with the elementwise sigmoid function $\bm{\sigma}$,
and $\odot$ denotes elementwise multiplication.
$\mW_u$ and $\mW_f$ are learned matrices.
Bias terms are suppressed for clarity.\footnote{Unlike other  RNNs
  such as LSTM \cite{Hochreiter:1997} or GRU \cite{Cho:2014}, RRNNs do
  not apply an affine transformation or a nonlinear dependency
	of $\vc_t$ on $\vc_{t-1}$.}

Slightly overloading the notation,
let $\overrightarrow{\vc}_{i,j}$ denote the encoding 
of span $[i,j]$ by running a forward RRNN in \equref{rrnn},
from left to right.
It can be efficiently computed
by subtracting $\overrightarrow{\vc}_{i-1}$ from $\overrightarrow{\vc}_{j}$,
weighted by a product of forget gates:
\begin{align}\label{eq:span}
\overrightarrow{\vc}_{i,j} = \overrightarrow{\vc}_j - \overrightarrow{\vc}_{i-1}\bigodot_{k=i}^{j}\overrightarrow{\vf}_{k}.
\end{align}
$\overrightarrow{\vf}_{k}$ vectors are the forget gates.
See~\S\ref{app:derivation} for a detailed derivation.
	
Using this observation,
we now derive an efficient algorithm to
calculate the span representations based on 
bidirectional RRNNs.
In the interest of space, \algoref{span_repr}
describes the forward span representations.
It takes advantage of the distributivity 
property of rational RNNs~\citep{Peng:2018},
and the number of RNN function calls
is linear in the input length.\footnote{In contrast, the dynamic
  program of~\citet{kong2015segrnn} for segmental (span-scoring) RNNs
	requires a quadratic number of recurrent function calls,
	since they use LSTMs, where distributivity does not hold.}
Although overall asymptotic time complexity is still 
quadratic,~\algoref{span_repr} only involves
elementwise operations, which can be
easily parallelized on modern GPUs.
The backward one (right to left) is analogous.

\begin{algorithm}[!t]
	\centering
	\small
	\caption{RRNN-based span representation.\footnotemark}
	\label{algo:span_repr}
	\begin{algorithmic}[1]
		\Procedure{SpanRepr}{$\{\protect\overrightarrow{\vc}_t, \protect\overrightarrow{\vf}_t\}$}
		\State\Comment Accumulate forward forget gates
		\For{$i=1, \dots, n$}
		\State $\overrightarrow{\vf}_{1,i}= \overrightarrow{\vf}_{1,i-1}\odot \overrightarrow{\vf}_{i}$
		\label{line:fw_f}
		\EndFor
		
		\For {$j=1,\dots, n$}
		\For{${i=1, \dots,j}$}
		\State$\overrightarrow{\vc}_{i,j}\hspace{-.1cm}=\hspace{-.1cm} \overrightarrow{\vc}_{j}- \overrightarrow{\vc}_{i-1}\odot\overrightarrow{\vf}_{1,j} / \overrightarrow{\vf}_{1,i-1}$
		\EndFor
		\EndFor
		\State\Return $\overrightarrow{\vc}_{i,j}$ vectors 
		\EndProcedure
	\end{algorithmic}
\end{algorithm}

\paragraph{Computing attention.}
As in standard attention,
we use a normalized weighted sum of the span representations. 
Let $\vg([i,j])=[\overrightarrow{\vc}_{i,j}; \overleftarrow{\vc}_{i,j}]$ denote the representation of span $[i,j]$,
which concatenates the forward and backward representations calculated using
\algoref{span_repr}.
The context vector $\va_t$ is
\begin{subequations}
\begin{align}
\va_{t+1} &= \sum_{i=0}^{m-1}\omega_{t,i}\;\vg([t-i,t]),\label{eq:context}\\
\omega_{t,i} &= \frac{\exp{s_{t,i}}}{\sum_{j=0}^{m-1}\exp{s_{t,j}}}.\label{eq:attention}
\end{align}
\end{subequations}
Here $s_{t,i}$ 
is implemented as an MLP,
taking as input the concatenation  
of $\vh_{t+1}$ and $\vg([t-i,t])$
and outputs the attention score.
The context vector is then concatenated
with the hidden state
$\bar{\vh}_{t+1}=[\vh_{t+1}; \va_{t+1}]$,
and fed into onward computation.

\footnotetext{$/$ denotes elementwise division.
	Both elementwise product and division are implemented in $\log$-space.}

In summary, given an input sequence, \modelname:
\begin{compactenum}
	\item First uses a standard left-to-right RNN to calculate
	the hidden states $\vh_t$.
	\item Feed $\vh_t$ vectors into a one-layer bidirectional
	rational RNN~(Eq.~\ref{eq:rrnn}), using \algoref{span_repr} to
	compute the span representations.
	\item Attends over spans~(Eq.~\ref{eq:attention}) to predict
          the next word.
\end{compactenum}

\isubsection{Attention-Based Constituency Parsing}{span_parsing}
We next describe the other facet or \modelname: the constituency
parser.
Our parsing algorithm is similar to the greedy top-down algorithm proposed by~\citet{Stern:2017}.
It recursively divides a span into two smaller ones,
until a single-token span, i.e., a leaf, is reached.
The order of the partition specifies 
the tree structure.\footnote{It is only able to produce binarized
unlabeled trees.}
Formally, for a maximum span length $m$, at each time step $j+1$, we split the span $[j-m+1, j]$
into two smaller parts $[j-m+1, k_0]$ and $[k_0+1, j]$.
The partitioning point is greedily selected, 
maximizing the attention scores of spans ending at $j$:\footnote{Another natural choice is 
	to maximize the sum of the scores of $[i, k_0]$ and $[k_0+1, j]$.
The attention score of $[i, k_0]$ is computed 
at time step $k_0$, and hence does not know
anything about the other span on the right.
Therefore we consider only the score of the right span.
}
\begin{align}
k_0 =\argmax{k \in \{0,\dots,m-1\}} s_{j,k}.
\end{align}
The span is directly returned as a leaf  
if it contains a single token.
A full parse is derived by running the algorithm recursively, 
starting with the input as a single span (with a special end-of-sentence mark at the end).
The runtime is $\gO(n^2)$,
with $n-1$ partitioning points.
See \figref{example} for an illustration.

\paragraph{Supervising the attention.}
Now that we are able to derive phrase structures from attention
weights, we can further inform the attention if syntactic
annotations are available, using oracle span selections.
For each token, the gold selection is a 
$m$-dimensional binary vector, and then
normalized to sum to one, denoted $\vy_t$.\footnote{Not necessarily one-hot:
	multiple spans can end at the same token. 
}
We add a cross-entropy loss (averaged across the training data) to the language modeling objective, 
with $\lambda$ trading off between the two:
\begin{align}\label{eq:loss}
\gL = \gL_{\text{LM}} + \frac{\lambda}{N}\sum_{t=1}^{N} \gH(\vy_t, \bm{\omega}_t),
\end{align}
with $\bm{\omega}$ being the attention distribution at step $t$, 
and $N$ the length of the training corpus.
As we will see in \secref{experiments}, 
providing syntactically guided span attention
improves language modeling performance.

\paragraph{Discussion.}
\modelname~provides an intuitive way to
inject structural inductive bias into the language model---by supervising the attention distribution.
This setting can be seen as a very lightweight 
multitask learning, where no actual syntactic tree
is predicted during language modeling training or evaluation.
The attention weight predictor (i.e., the $s$ scores in Eq.~\ref{eq:attention})
can be replaced with an off-the-shelf parser,
or deterministically set (e.g., to simulate left/right-branching).

%% file: text/experiments.tex
	\isection{Experiments}{experiments}
\label{sec:experiments}


We evaluate \modelname on language modeling.
We experiment with the Penn Treebank corpus~(PTB) and WikiText-2 (WT2).
We follow the preprocessing of \citet{mikolov2010rnn} for PTB and \citet{Merity:2018} for WT2. 
More implementation details are described in \appref{implementation}. 
We compare two configurations of \modelname:
\begin{compactenum}[$\bullet$]
	\item \modelname-U builds on top of AWD-LSTM~\citep{Merity:2018}, a state-of-the-art of
	LSTM implementation for language modeling. The span attention is included
        \emph{before} the last
 layer.\footnote{Preliminary experiments show that including the span attention after the last layer
		yields similar empirical results, but is more sensitive to hyperparameters.
		}
	\item \modelname-S is the same model as \modelname-U,
	but uses phrase syntax annotation to 
	provide additional supervision to the attention component~(\secref{span_parsing}).\footnote{We use the WSJ portion of PTB for parsing annotations.}
\end{compactenum}

We compare against the AWD-LSTM baseline.
On PTB, we also compare to two models using structural information in language modeling:
parsing-reading-predict networks~(PRPN;~\citealp{shen2018neural})
predicts syntactic distance as structural features for language modeling;
ordered-neuron LSTM~(ON-LSTM;~\citealp{shen2018ordered}) posits a novel ordering on LSTM gates,
simulating the covering of phrases at different levels
in a constituency parse.
On PTB we also compare to \modelname-RB,
a baseline deterministically setting the attention scores (Eq.~\ref{eq:attention}) in decreasing order,
such that the derived trees will be right-branching.\footnote{We set scores to $m,m-1,\dots,1$, before the $\operatorname{softmax}$.
}

Tables~\ref{table:lm} and~\ref{table:lm_wiki} summarize the language modeling results.
On both datasets, the unsupervised configuration (\modelname-U)
outperforms AWD-LSTM. 
On PTB, \modelname-U achieves similar performance to ON-LSTM and much better performance than PRPN.
\modelname-S further reduces the perplexity by 1.6--3.4\%  (relative),
showing that incorporating structural information with supervised span
attention helps language modeling.
Naively promoting right-branching attention (\modelname-RB) yields no improvement 
over the baseline.



\begin{table}[tb]
	\centering
	\begin{tabulary}{0.47\textwidth}{@{}l  c cc@{}} 
		
		\toprule
		
		\textbf{Model}
		& \textbf{\# Params.}
		& \textbf{Dev.}
		& \textbf{Test}\\
		
		\midrule
		AWD-LSTM & 24M & 60.0 & 57.3  \\
		\midrule[.03em]
		PRPN & - &- & 62.0  \\
		ON-LSTM  & 25M & 58.3 &56.2 \\
		\midrule[.03em]
		\modelname-U & 24M & 58.6 & 56.4  \\
		\modelname-RB & 24M & 60.1 & 57.5 \\
		\modelname-S & 24M & \textbf{57.9} & \textbf{55.5} \\

		\bottomrule
	\end{tabulary}
	\caption{PTB language modeling perplexity (lower is better).
		Bold fonts indicate best performance.\footnotemark}
	\label{table:lm}
\end{table}
\footnotetext{Several recent works report better
	language modeling perplexity~\interalia{yang2017breaking,takase2018direct,dai2019transformer}.
	Their contribution is orthogonal to ours and not head-to-head comparable to the models in the table.
	}

\begin{table}[tb]
	\centering
	\begin{tabulary}{0.47\textwidth}{@{}l  c cc@{}} 
		
		\toprule
		
		\textbf{Model}
		& \textbf{\# Params.}
		& \textbf{Dev.}
		& \textbf{Test}\\
		
		\midrule
		AWD-LSTM & 33M & 68.6 & 65.8 \\
		\midrule[.03em]
		\modelname-U & 36M & 68.4 & 65.4\\
		\modelname-S & 36M & \textbf{65.5} & \textbf{63.2} \\
		\bottomrule
	\end{tabulary}
	\caption{WikiText-2 language modeling perplexity (lower is better).
		Bold fonts indicate best performance.}
	\label{table:lm_wiki}
\end{table}

\paragraph{Unsupervised constituency parsing.}
We evaluate the parser component of \modelname-U on WSJ-40.
It uses the same data as in language modeling,
but filters out sentences longer than 40 tokens
after punctuation removal.
The model is selected based on language modeling validation perplexity.

In addition to PRPN, we compare to DIORA~\citep{Drozdov:2019},
which uses an inside-outside dynamic program 
in an autoencoder.
Table~\ref{table:parsing} shows the $F_1$ results.
\modelname outperforms the right branching baseline,
but is not as accurate as the other models.\footnote{
	Evaluation on WSJ-10, which contains sentences with 10 or less tokens,
	 shows a similar trend.}\com{ is used by several recent works~\citep{shen2018neural,shen2018ordered,htut2018grammar}.
	It reveals some test sentences (but \emph{not} labels) to their models during		
	training.		
	Right branching baseline achieves 61.7 F$_{1}$ under this setting;		
	PRPN: 70.5; ON-LSTM: 65.1; \modelname-U: 62.9.
	\citet{kim2019unsupervised} uses considerably 
	different step ups,
	and is not comparable to ours.} 
This indicates that the type of syntactic trees learned by it, albeit useful to the LM component, 
do not correspond well to PTB-like syntactic trees.


\paragraph{Discussion.}
Despite its strong performance,
the parsing algorithm used by
\citet{shen2018neural} and \citet{shen2018ordered}
suffers from an incomplete support issue.\footnote{Chris Dyer, personal communication.}
More precisely, 
it fails to produce ``close-open-open,'' i.e., \texttt{)((} structures.
As a result,
the parser is intrinsically biased toward right-branching structures.
\modelname, on the other hand, scores all the spans,
and therefore can produce any binary tree spanning a given sentence:
the algorithm recovers any given binary tree by 
letting $s_{j,j-i} =1$ if the tree contains nonterminal $[i, j]$,
and 0 otherwise.\footnote{
	The maximum span length $m$ is only forced in language modeling training and evaluation.}

Is \modelname~empirically biased toward any branching direction?
In greedily selected trees, we measure the percentage of left-branching splits (dividing $[i,j]$ into $[i, j-1]$ and $j$)
and right-branching splits (dividing $[i,j]$ into $i$ and $[i+1, j]$).%
\footnote{We exclude trivial splits dividing a length-2 span into two tokens.}
Table~\ref{table:branching} summarizes the results on WSJ-40 test set.
The first row shows the results for randomly initialized models without training.
We observe no significant trend of favoring one branching direction over the other.
However, after training with the language modeling objective,
\modelname-U shows a clear right-skewness more than it should: 
it produces much more right-branching structures
than the gold annotation.  This means that the
span attention mechanism has learned to emphasize longer prefixes, 
rather than make strong Markov assumptions.
More exploration of this effect is left to future work.

\begin{table}[tb]
	\centering
	\begin{tabulary}{0.47\textwidth}{@{}l  c @{}} 
		
		\toprule
		
		\textbf{Model}
		& \textbf{Unlabeled $F_1$}\\
		
		\midrule
		Right Branching 
		& 40.7 \\
		\midrule[.03em]

		$^\dagger$DIORA 
		& 60.6 \\
		$^\ddagger$PRPN
		& 52.4 \\
		\midrule[.03em]

		$^\ddagger$\modelname-U
		& 42.0 \\
		\bottomrule
	\end{tabulary}
	\caption{Unlabeled unsupervised parsing $F_1$ on WSJ-40.
		$\ddagger$ trains on the training split of WSJ, while $\dagger$ trains on
		AllNLI~\citep{htut2018grammar}.
		The PRPN result is taken from~\citet{Drozdov:2019}.
		}
	\label{table:parsing}
\end{table}

\begin{table}[tb]
	\centering
	\begin{tabulary}{0.47\textwidth}{@{}l  rrrr@{}} 
		
		\toprule

		& \multicolumn{2}{c}{\textbf{\% Left Splits}}
		& \multicolumn{2}{c}{\textbf{\% Right Splits}}\\
		
		\midrule
		Random & 39.3&\small ${\pm10.5}$ & 41.2& \small ${\pm8.8}$  \\
		\modelname-U & 1.1 && 85.6& \\
		\midrule[.03em]
		Gold & 6.5 && 52.7& \\
		\bottomrule
	\end{tabulary}
	\caption{Percentage of left and right splits. The first row shows the numbers 
		averaging over 25 differently randomly initialized \modelname models, without training.
		$\pm$ indicates standard deviation.}
	\label{table:branching}
\end{table}


%% file: text/conclusion.tex
\section{Conclusion}
We present~\modelname, a hybrid parser and language model.
\modelname attends over the preceding text spans.
From its attention weights phrase structures can be derived.
The attention component can be separately trained
to provide syntactically-informed context gathering.
\modelname outperforms
strong baselines on language modeling.
Incorporating syntactic supervision during training leads to further
language modeling improvements.
Training our unsupervised model on large-scale corpora could
result in both stronger language models and, potentially, stronger parsers.
Our code is publicly available at \repo.

\section*{Acknowledgments}
We thank members of the ARK at the University of Washington,
and researchers at the Allen Institute for Artificial Intelligence 
for their helpful comments on an earlier version of this work,
and the anonymous reviewers for their insightful feedback.
This work was supported in part by NSF grant 1562364.

%% file: text/appendix.tex
\begin{appendices}
	\section{Implementation Details}\label{app:implementation}
	\paragraph{Neural Network Architecture}
	Our implementation is based on AWD-LSTM~\citep{Merity:2018}.\footnote{\url{https://github.com/salesforce/awd-lstm-lm}}
	It uses a three-layer LSTM, with carefully designed regularization techniques.
	\modelname includes the span attention after the second layer.
	Preliminary results show that 
	it yields similar results, but is less sensitive to
	hyperparameters, compared to adding it to the last layer.
	
	The context is concatenated to the hidden state ($\bar{\vh}_{t}=[\vh_{t}; \va_{t}]$),
	and then fed to a $\tanh$-MLP controlled by a residual gate $\vg_r$~\citep{he2016deep}, 
	before fed onward into the next LSTM layer:
	\begin{align}
		\hat{\vh}_t = \vg_r \odot \operatorname{MLP}\left(\bar{\vh}_t\right) + 
		(\bm{1} -\vg_r)\odot\vh_t.
	\end{align}
	The rest of the architecture stays the same as AWD-LSTM. 
	We refer the readers to \citet{Merity:2018} for more details.
	
	\paragraph{More details on \modelname-S.}
	\modelname-S uses exactly the same architecture and hyperparameters as
	its unsupervised counterpart.
	We derive, from PTB training data, a $m$-dimensional 0-1 vector for each token.
	Each element specifies whether the corresponding 
	span appears in the gold parse.
	Trivial spans (i.e., the ones over single tokens and full sentences)
	are ignored.
	The vector are normalized to sum to one,
	in order to facilitate the use of cross-entropy loss.
	$\lambda$ in Eq.~\ref{eq:loss} is set to 0.01.
	
	\paragraph{Hyperparameters.}
	The regularization and hyperparameters largely follow \citet{Merity:2018}.
	We only differ from them by using smaller hidden size (and hence smaller dropout rate)
	to control for the amount of parameters in the PTB
        experiments, summarized in Table~\ref{tab:hyperparameters} 
	For the WikiText-2 experiments, we use 200 rational RNN size and 400 dimensional context vectors.
	Other hyperparameters follow \citet{Merity:2018}.
	The max span length $m$ is set to 20 for PTB experiments,
	and 10 for WikiText-2. 
	
	\citet{Merity:2018} start by using SGD to train the model,
	and switch to averaged SGD~\citep{polyak1992asgd}
	after 5 nonimprovement-epochs.
	We instead use Adam~\citep{kingma2014adam} 
	with default PyTorch settings
	to train the model for 40 epochs, and then switch to ASGD,
	allowing for faster convergence.
	
	\begin{table}[tb]
		\centering
		\begin{tabular}{lr}
			\toprule
			{\bf Type} & {\bf Values}  \\
			\midrule
			Rational RNN size & 200 \\
			Context Vector Size & 400 \\
			\midrule
			LSTM Hidden Size & 1020  \\  
			Weight Dropout & 0.45 \\ 
			Vertical Dropout &  0.2 \\ 
			\bottomrule
		\end{tabular}
		\caption{\label{tab:hyperparameters} 
			The hyperparameters used in the PTB language modeling experiment.}
	\end{table}
	
	\section{Span Representations}\label{app:derivation}
	Below is the derivation for Eq.~\ref{eq:span}.
	
	\begin{align*}
	\overrightarrow{\vc}_{i,j}
	&= \overrightarrow{\vu}_{j} + \sum_{k=i}^{j-1}\overrightarrow{\vu}_{k}\bigodot_{\ell=k+1}^{j}\overrightarrow{\vf}_\ell\\
	&=\overrightarrow{\vu}_{j} + \sum_{k=1}^{j-1}\overrightarrow{\vu}_{k}\bigodot_{\ell=k+1}^{j}\overrightarrow{\vf}_\ell
		-\sum_{k=1}^{i-1}\overrightarrow{\vu}_{k}\bigodot_{\ell=k+1}^{j}\overrightarrow{\vf}_\ell\\
	&= \overrightarrow{\vc}_j 
	- \left(
	\overrightarrow{\vu}_{i-1} 
	+ \sum_{k=1}^{i-2}
	\overrightarrow{\vu}_{k}\bigodot_{\ell=k+1}^{i-1}\overrightarrow{\vf}_\ell
	\right)\bigodot_{\ell=i}^{j}\overrightarrow{\vf}_\ell\\
	&= \overrightarrow{\vc}_j -\overrightarrow{\vc}_{i-1} \bigodot_{k=i}^{j}\overrightarrow{\vf}_k
	\end{align*}

\end{appendices}

%% file: syntactic_lm.bbl
\begin{thebibliography}{40}
\expandafter\ifx\csname natexlab\endcsname\relax\def\natexlab#1{#1}\fi

\bibitem[{Buys and Blunsom(2018)}]{buys2018neural}
Jan Buys and Phil Blunsom. 2018.
\newblock Neural syntactic generative models with exact marginalization.
\newblock In \emph{Proc. of NAACL}.

\bibitem[{Chelba and Jelinek(1998)}]{chelba1998exploiting}
Ciprian Chelba and Frederick Jelinek. 1998.
\newblock Exploiting syntactic structure for language modeling.
\newblock In \emph{Proc. of {COLING}}.

\bibitem[{Cho et~al.(2014)Cho, Van~Merri{\"e}nboer, Gulcehre, Bahdanau,
  Bougares, Schwenk, and Bengio}]{Cho:2014}
Kyunghyun Cho, Bart Van~Merri{\"e}nboer, Caglar Gulcehre, Dzmitry Bahdanau,
  Fethi Bougares, Holger Schwenk, and Yoshua Bengio. 2014.
\newblock Learning phrase representations using {RNN} encoder-decoder for
  statistical machine translation.
\newblock In \emph{Proc. of EMNLP}.

\bibitem[{Dai et~al.(2019)Dai, Yang, Yang, Carbonell, Le, and
  Salakhutdinov}]{dai2019transformer}
Zihang Dai, Zhilin Yang, Yiming Yang, Jaime~G. Carbonell, Quoc~V. Le, and
  Ruslan Salakhutdinov. 2019.
\newblock {Transformer-XL}: Attentive language models beyond a fixed-length
  context.
\newblock In \emph{Proc. of ACL}.

\bibitem[{Das et~al.(2012)Das, Martins, and Smith}]{Das:2012}
Dipanjan Das, Andr{\'e} F.~T. Martins, and Noah~A. Smith. 2012.
\newblock An exact dual decomposition algorithm for shallow semantic parsing
  with constraints.
\newblock In \emph{Proc. of *{SEM}}.

\bibitem[{Drozdov et~al.(2019)Drozdov, Verga, Yadav, Iyyer, and
  McCallum}]{Drozdov:2019}
Andrew Drozdov, Pat Verga, Mohit Yadav, Mohit Iyyer, and Andrew McCallum. 2019.
\newblock Unsupervised latent tree induction with deep inside-outside recursive
  autoencoders.
\newblock In \emph{Proc. of NAACL}.

\bibitem[{Dyer et~al.(2016)Dyer, Kuncoro, Ballesteros, and
  Smith}]{dyer2016rnng}
Chris Dyer, Adhiguna Kuncoro, Miguel Ballesteros, and Noah~A. Smith. 2016.
\newblock Recurrent neural network grammars.
\newblock In \emph{Proc. of NAACL}.

\bibitem[{Gildea and Palmer(2002)}]{Gildea:2002}
Daniel Gildea and Martha Palmer. 2002.
\newblock The necessity of parsing for predicate argument recognition.
\newblock In \emph{Proc. of ACL}.

\bibitem[{He et~al.(2016)He, Zhang, Ren, and Sun}]{he2016deep}
Kaiming He, Xiangyu Zhang, Shaoqing Ren, and Jian Sun. 2016.
\newblock Deep residual learning for image recognition.
\newblock \emph{Proc. of CVPR}.

\bibitem[{Hochreiter and Schmidhuber(1997)}]{Hochreiter:1997}
Sepp Hochreiter and J{\"u}rgen Schmidhuber. 1997.
\newblock Long short-term memory.
\newblock \emph{Neural computation}, 9(8):1735--1780.

\bibitem[{Howard and Ruder(2018)}]{Howard:2018}
Jeremy Howard and Sebastian Ruder. 2018.
\newblock Universal language model fine-tuning for text classification.
\newblock In \emph{Proc. of ACL}.

\bibitem[{Htut et~al.(2018)Htut, Cho, and Bowman}]{htut2018grammar}
Phu~Mon Htut, Kyunghyun Cho, and Samuel Bowman. 2018.
\newblock Grammar induction with neural language models: An unusual
  replication.
\newblock In \emph{Proc. of EMNLP}.

\bibitem[{Jelinek and Lafferty(1991)}]{jelinek1991computation}
Frederick Jelinek and John~D. Lafferty. 1991.
\newblock Computation of the probability of initial substring generation by
  stochastic context-free grammars.
\newblock \emph{Computational Linguistics}, 17(3):315--353.

\bibitem[{Jozefowicz et~al.(2016)Jozefowicz, Vinyals, Schuster, Shazeer, and
  Wu}]{Jozefowicz:2016}
Rafal Jozefowicz, Oriol Vinyals, Mike Schuster, Noam Shazeer, and Yonghui Wu.
  2016.
\newblock Exploring the limits of language modeling.
\newblock {arXiv}:1602.02410.

\bibitem[{Kim et~al.(2019)Kim, Rush, Yu, Kuncoro, Dyer, and
  Melis}]{kim2019unsupervised}
Yoon Kim, Alexander~M. Rush, Lei Yu, Adhiguna Kuncoro, Chris Dyer, and
  G{\'{a}}bor Melis. 2019.
\newblock Unsupervised recurrent neural network grammars.
\newblock In \emph{Proc. of NAACL}.

\bibitem[{Kingma and Ba(2014)}]{kingma2014adam}
Diederik~P. Kingma and Jimmy Ba. 2014.
\newblock Adam: {A} method for stochastic optimization.
\newblock {arXiv}:1412.6980.

\bibitem[{Kong et~al.(2016)Kong, Dyer, and Smith}]{kong2015segrnn}
Lingpeng Kong, Chris Dyer, and Noah~A. Smith. 2016.
\newblock Segmental recurrent neural networks.
\newblock In \emph{Proc. of ICLR}.

\bibitem[{Lee et~al.(2017)Lee, He, Lewis, and Zettlemoyer}]{Lee:2017}
Kenton Lee, Luheng He, Mike Lewis, and Luke Zettlemoyer. 2017.
\newblock End-to-end neural coreference resolution.
\newblock In \emph{Proc. of EMNLP}.

\bibitem[{Marcus et~al.(1993)Marcus, Marcinkiewicz, and
  Santorini}]{Marcus:1993}
Mitchell~P. Marcus, Mary~Ann Marcinkiewicz, and Beatrice Santorini. 1993.
\newblock Building a large annotated corpus of {E}nglish: {T}he {P}enn
  {T}reebank.
\newblock \emph{Computational Linguistics}, 19(2):313--330.

\bibitem[{Merity et~al.(2018)Merity, Keskar, and Socher}]{Merity:2018}
Stephen Merity, Nitish~Shirish Keskar, and Richard Socher. 2018.
\newblock Regularizing and optimizing {LSTM} language models.
\newblock In \emph{Proc. of ICLR}.

\bibitem[{Merity et~al.(2017)Merity, Xiong, Bradbury, and Socher}]{Merity:2017}
Stephen Merity, Caiming Xiong, James Bradbury, and Richard Socher. 2017.
\newblock Pointer sentinel mixture models.
\newblock In \emph{Proc. of ICLR}.

\bibitem[{Mikolov et~al.(2010)Mikolov, Karafi{\'a}t, Burget,
  {\v{C}}ernock{\`y}, and Khudanpur}]{mikolov2010rnn}
Tom{\'a}{\v{s}} Mikolov, Martin Karafi{\'a}t, Luk{\'a}{\v{s}} Burget, Jan
  {\v{C}}ernock{\`y}, and Sanjeev Khudanpur. 2010.
\newblock Recurrent neural network based language model.
\newblock In \emph{Proc. of INTERSPEECH}.

\bibitem[{Parikh et~al.(2016)Parikh, T{\"a}ckstr{\"o}m, Das, and
  Uszkoreit}]{parikh2016decomposable}
Ankur Parikh, Oscar T{\"a}ckstr{\"o}m, Dipanjan Das, and Jakob Uszkoreit. 2016.
\newblock A decomposable attention model for natural language inference.
\newblock In \emph{Proc. of EMNLP}.

\bibitem[{Peng et~al.(2018{\natexlab{a}})Peng, Schwartz, Thomson, and
  Smith}]{Peng:2018}
Hao Peng, Roy Schwartz, Sam Thomson, and Noah~A. Smith. 2018{\natexlab{a}}.
\newblock Rational recurrences.
\newblock In \emph{Proc. of EMNLP}.

\bibitem[{Peng et~al.(2018{\natexlab{b}})Peng, Thomson, and
  Smith}]{peng2018backprop}
Hao Peng, Sam Thomson, and Noah~A. Smith. 2018{\natexlab{b}}.
\newblock Backpropagating through structured argmax using a spigot.
\newblock In \emph{Proc. of ACL}.

\bibitem[{Peng et~al.(2018{\natexlab{c}})Peng, Thomson, Swayamdipta, and
  Smith}]{peng2018learning}
Hao Peng, Sam Thomson, Swabha Swayamdipta, and Noah~A. Smith.
  2018{\natexlab{c}}.
\newblock Learning joint semantic parsers from disjoint data.
\newblock In \emph{Proc. of NAACL}.

\bibitem[{Peters et~al.(2018)Peters, Neumann, Iyyer, Gardner, Clark, Lee, and
  Zettlemoyer}]{Peters:2018}
Matthew Peters, Mark Neumann, Mohit Iyyer, Matt Gardner, Christopher Clark,
  Kenton Lee, and Luke Zettlemoyer. 2018.
\newblock Deep contextualized word representations.
\newblock In \emph{Proc. of NAACL}.

\bibitem[{Polyak and Juditsky(1992)}]{polyak1992asgd}
B.~T. Polyak and A.~B. Juditsky. 1992.
\newblock Acceleration of stochastic approximation by averaging.
\newblock \emph{SIAM J. Control Optim.}, 30(4):838--855.

\bibitem[{Punyakanok et~al.(2008)Punyakanok, Roth, and Yih}]{Punyakanok:2008}
Vasin Punyakanok, Dan Roth, and Wen-tau Yih. 2008.
\newblock The importance of syntactic parsing and inference in semantic role
  labeling.
\newblock \emph{American Journal of Computational Linguistics}, 34(2):257--287.

\bibitem[{Radford et~al.(2019)Radford, Wu, Child, Luan, Amodei, and
  Sutskever}]{Radford:2019}
Alec Radford, Jeffrey Wu, Rewon Child, David Luan, Dario Amodei, and Ilya
  Sutskever. 2019.
\newblock Language models are unsupervised multitask learners.
\newblock OpenAI Blog.

\bibitem[{Roark(2001)}]{roark2001probabilistic}
Brian Roark. 2001.
\newblock Probabilistic top-down parsing and language modeling.
\newblock \emph{Computational Linguistics}, 27(2):249--276.

\bibitem[{Shen et~al.(2018{\natexlab{a}})Shen, Lin, wei Huang, and
  Courville}]{shen2018neural}
Yikang Shen, Zhouhan Lin, Chin wei Huang, and Aaron Courville.
  2018{\natexlab{a}}.
\newblock Neural language modeling by jointly learning syntax and lexicon.
\newblock In \emph{Proc. of ICLR}.

\bibitem[{Shen et~al.(2018{\natexlab{b}})Shen, Tan, Sordoni, and
  Courville}]{shen2018ordered}
Yikang Shen, Shawn Tan, Alessandro Sordoni, and Aaron~C. Courville.
  2018{\natexlab{b}}.
\newblock Ordered neurons: Integrating tree structures into recurrent neural
  networks.
\newblock In \emph{Proc. of ICLR}.

\bibitem[{Stern et~al.(2017)Stern, Andreas, and Klein}]{Stern:2017}
Mitchell Stern, Jacob Andreas, and Dan Klein. 2017.
\newblock A minimal span-based neural constituency parser.
\newblock In \emph{Proc. of ACL}.

\bibitem[{Strubell et~al.(2018)Strubell, Verga, Andor, Weiss, and
  McCallum}]{Strubell:2018}
Emma Strubell, Patrick Verga, Daniel Andor, David Weiss, and Andrew McCallum.
  2018.
\newblock Linguistically-informed self-attention for semantic role labeling.
\newblock In \emph{Proc. of EMNLP}.

\bibitem[{Swayamdipta et~al.(2018)Swayamdipta, Thomson, Lee, Zettlemoyer, Dyer,
  and Smith}]{Swayamdipta:2018}
Swabha Swayamdipta, Sam Thomson, Kenton Lee, Luke Zettlemoyer, Chris Dyer, and
  Noah~A. Smith. 2018.
\newblock Syntactic scaffolds for semantic structures.
\newblock In \emph{Proc. of EMNLP}.

\bibitem[{Takase et~al.(2018)Takase, Suzuki, and Nagata}]{takase2018direct}
Sho Takase, Jun Suzuki, and Masaaki Nagata. 2018.
\newblock Direct output connection for a high-rank language model.
\newblock In \emph{Proc. of EMNLP}.

\bibitem[{Vaswani et~al.(2017)Vaswani, Shazeer, Parmar, Uszkoreit, Jones,
  Gomez, Kaiser, and Polosukhin}]{vaswani2017attention}
Ashish Vaswani, Noam Shazeer, Niki Parmar, Jakob Uszkoreit, Llion Jones,
  Aidan~N Gomez, \L~ukasz Kaiser, and Illia Polosukhin. 2017.
\newblock Attention is all you need.
\newblock In \emph{Proc. of NeurIPS}.

\bibitem[{Yang et~al.(2019)Yang, Dai, Salakhutdinov, and
  Cohen}]{yang2017breaking}
Zhilin Yang, Zihang Dai, Ruslan Salakhutdinov, and William~W. Cohen. 2019.
\newblock Breaking the softmax bottleneck: {A} high-rank {RNN} language model.
\newblock In \emph{Proc. of ICLR}.

\bibitem[{Zellers et~al.(2018)Zellers, Bisk, Schwartz, and Choi}]{Zellers:2018}
Rowan Zellers, Yonatan Bisk, Roy Schwartz, and Yejin Choi. 2018.
\newblock {SWAG}: A large-scale adversarial dataset for grounded commonsense
  inference.
\newblock In \emph{Proc. of EMNLP}.

\end{thebibliography}
